\pdfoutput=1

\documentclass[11pt]{article}

\usepackage{acl}

\usepackage{times}
\usepackage{latexsym}

\usepackage[T1]{fontenc}

\usepackage[utf8]{inputenc}

\usepackage{microtype}
\usepackage{times}
\usepackage{latexsym}
\usepackage{booktabs} 
\usepackage{multirow}
\usepackage{url}
\usepackage{amsmath}
\usepackage{bm}
\usepackage{amsfonts}
\usepackage{graphicx}
\usepackage{diagbox}
\usepackage{xspace}
\usepackage{courier}
\usepackage{mathrsfs}
\usepackage{appendix}
\usepackage{float}
\usepackage{framed} 
\usepackage{color}
\usepackage{xspace}
\usepackage{amsfonts}
\usepackage{amssymb}
\usepackage{diagbox}

\usepackage{amsmath}

\usepackage{microtype}
\usepackage{textcomp}
\usepackage{amsfonts}
\usepackage{amssymb}
\usepackage{diagbox}
\usepackage{algorithm}
\usepackage{algorithmic}

\newcommand\Ourmodel{PELT\xspace}

\title{A Simple but Effective Pluggable Entity Lookup Table \\ for Pre-trained Language Models}

 \author{Deming Ye$^{1,2}$, Yankai Lin$^{6}$, Peng Li$^{6,7}$,  Maosong Sun$^{1,2,3,4,5}$\thanks{ \ \ Corresponding author: M. Sun (sms@tsinghua.edu.cn)}\,\,,  Zhiyuan Liu$^{1,2,3,5}$ \\
$^{1}$Dept. of Comp. Sci. \& Tech., Institute for AI, Tsinghua University, Beijing, China \\
$^{2}$Beijing National Research Center for Information Science and Technology \\
$^{3}$International Innovation Center of Tsinghua University, Shanghai, China \\
$^{4}$Jiangsu Collaborative Innovation Center for Language Ability, Xuzhou, China \\
$^{5}$Institute Guo Qiang, Tsinghua University $^{6}$Pattern Recognition Center, WeChat AI \\
$^{7}$Institute for AI Industry Research (AIR), Tsinghua University\\
\texttt{yedeming001@163.com}
}

\begin{document}
\maketitle
\begin{abstract}
Pre-trained language models (PLMs) cannot well recall rich factual knowledge of entities exhibited in large-scale corpora, especially those rare entities. In this paper, we propose to build a simple but effective \textbf{P}luggable \textbf{E}ntity \textbf{L}ookup \textbf{T}able (\Ourmodel) on demand by aggregating the entity's output representations of multiple occurrences in the corpora.  \Ourmodel can be compatibly plugged as inputs  to infuse supplemental entity knowledge  into PLMs. 
Compared to previous knowledge-enhanced PLMs,  \Ourmodel only requires 0.2\%$\sim$5\% pre-computation with capability of acquiring knowledge from  out-of-domain corpora for domain adaptation scenario. 
The experiments on knowledge-related tasks demonstrate that our method, PELT, can flexibly and effectively transfer entity knowledge from related corpora into PLMs with different architectures. Our code and models are publicly available at \url{https://github.com/thunlp/PELT}.

 
%
%




\end{abstract}

\section{Introduction}
Recent advance in pre-trained language models (PLMs) has achieved promising improvements in various downstream tasks~\cite{BERT,RoBERTa}. Some latest works reveal that PLMs can automatically acquire knowledge from large-scale corpora via self-supervised pre-training and then encode the  learned  knowledge into their model parameters~\cite{Prob, LAMA, T5closedbookqa}. 
However,  due to the limited capacity of vocabulary, existing PLMs face the challenge of recalling the factual knowledge from their parameters, especially for those rare entities~\cite{RepDegeneration,canclosedbookqa}.


To improve  PLMs' capability of entity understanding, a straightforward solution is to exploit an external entity embedding acquired from the knowledge graph (KG)~\cite{ERNIE, K-Adapter}, the  entity description~\cite{KnowBERT}, or the corpora~\cite{LAMA-UHN}. In order to make use of the external knowledge, these models usually learn to align the external entity embedding~\cite{TransE, Wikipedia2Vec} to the their original word embedding. However, previous works ignore to explore entity embedding from the PLM itself, which makes their learned embedding mapping is not available in the domain-adaptation.  
Other recent works attempt to infuse knowledge into PLMs' parameters by extra pre-training, such as learning to build an additional entity vocabulary from the corpora~\cite{LUKE,EAE}, or adopting  entity-related pre-training tasks to intensify the entity representation~\cite{ WKLM,CoLAKE, KEPLER}. However, their huge pre-computation increases the cost of extending or updating the customized vocabulary for various downstream tasks.



\begin{table}[t!]
\small
\centering
\begin{tabular}{l|c|r|c}
\toprule
\textbf{Model} & \textbf{\#Ent} &\multicolumn{1}{c|}{\textbf{Pre-Comp.}}  & \textbf{D-Adapt} \\ 
\midrule
\citet{ERNIE} & 5.0M &  $\sim$160h &  No\\ 
\citet{KEPLER} & 4.6M & $\sim$3,400h & No\\
\citet{LUKE} & 0.5M   & $\sim$3,800h & No\\
PELT (our model) & 4.6M & 7h &Yes\\
\bottomrule
\end{tabular}
 \caption{Comparison of recent knowledge-enhanced PLMs. We report the pre-computation of  \texttt{BASE} models on Wikipedia entities on a  V100 GPU. Pre-Comp.: Pre-computation; D-Adapt: Domain Adaptation.}
\vspace{-0.7em}
 \label{tab:computation}
\end{table}

In this paper, we introduce a simple but effective  \textbf{P}luggable \textbf{E}ntity \textbf{L}ookup \textbf{T}able (\Ourmodel) to infuse knowledge into PLMs. To be specific, we first  revisit the connection between PLMs' input features and output representations for masked language modeling. Based on this, given a new corpus, we aggregate the output representations of masked tokens from the entity's  occurrences, to recover an elaborate entity embedding from a well-trained PLM. Benefiting from the compatibility and flexibility of the constructed embedding, we can directly insert them into the corresponding positions of the input sequence to provide supplemental entity knowledge. As shown in Table~\ref{tab:computation},  our method merely consumes 0.2\%$\sim$5\% pre-computation compared with previous works, and it also supports the vocabulary from different domains simultaneously.

We conduct experiments on two knowledge-related tasks, including knowledge probe and  relation classification, across two domains (Wikipedia and biomedical publication).  Experimental results show that PLMs with \Ourmodel can consistently and significantly outperform the corresponding vanilla models. In addition, the entity embedding obtained from multiple domains are compatible with the original word embedding and can be applied and  transferred swiftly.

\section{Methodology}

In this section, we first revisit the masked language modeling pre-training objective. After that, we introduce the pluggable entity lookup table and explain how to apply it to incorporate knowledge into PLMs.

\subsection{Revisit Masked Language Modeling}
PLMs conduct self-supervised pre-training tasks, such as masked language modeling (MLM)~\cite{BERT}, to learn the semantic and syntactic knowledge from the large-scale unlabeled corpora~\cite{BERTology}.  
MLM can be regarded as a kind of cloze task, which requires the model to predict the missing tokens based on its contextual representation. Formally, given a sequence of tokens $X =(x_1, x_2, \ldots, x_n)$, with $x_i$ substituted by \texttt{[MASK]}, PLMs, such as BERT, first take tokens' word embedding and position embedding as input and obtain the contextual representation:
\begin{equation}
\bm{H} = \text{Enc}(\text{LayerNorm}({\textbf{E}}(X) + \bm{P})),
\label{eq1}
\end{equation}
where Enc$(\cdot)$ denotes a deep bidirectional Transformer encoder, LayerNorm$(\cdot)$ denotes layer normalization~\cite{LayerNorm},  ${\textbf{E}} \in {\mathbb{R}}^{|V| \times D}$ is the word embedding matrix, $V$ is the word vocabulary, ${P}$ is the absolute position embedding and $\bm{H} =(\bm{h}_1, \bm{h}_2, \ldots, \bm{h}_n)$ is the contextual representation. After that, BERT applies a feed-forward network (FFN)  and layer normalization on the contextual representation to compute the output representation of $x_i$:  
\begin{equation}
\bm{r}_{x_i} = \text{LayerNorm}( \text{FFN}(\bm{h}_i )).
\end{equation}

Since  the weights in the softmax layer and word embeddings are tied in BERT, the model calculate the product of $\bm{r}_{x_i}$ and the input word embedding matrix to further compute $x_i$'s cross-entropy loss among all the words:
\begin{equation}
\begin{split}
\mathcal{L} &= - \sum \log \text{Pr}(x_i|\bm{r}_{x_i}) \\
&= -\sum \log \frac{\exp({{\textbf{E}}(x_i)}^T\bm{r}_{x_i} )}{ \sum_{w_j \in V}{\exp({\textbf{E}}(w_j)}^T \bm{r}_{x_i}) }.
\end{split}
\label{eq:mlm}
\end{equation}

\subsection{Construct Pluggable Entity Embedding}
\label{sec:construct}

\begin{figure}[!t]
    \centering
    \includegraphics[width=\linewidth]{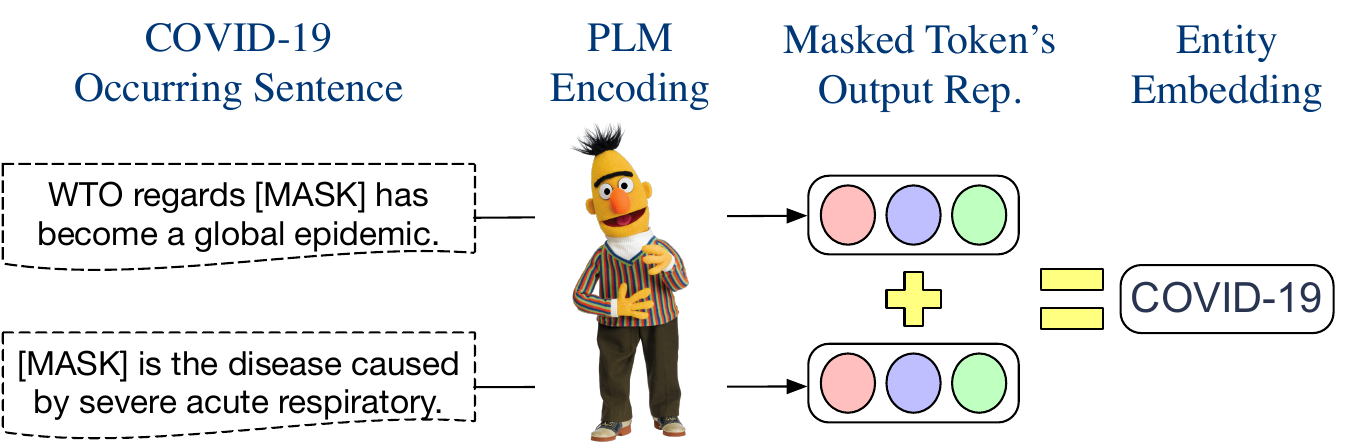}
    \caption{An illustration of the our PELT.}
    \label{fig:model}
\end{figure}

Due to the training efficiency, the vocabulary sizes in existing PLMs typically range from 30K to 60K subword units, and thus PLMs have to disperse the information of massive entities into their subword embeddings. Through revisiting the MLM loss in Eq.~\ref{eq:mlm}, we could intuitively observe that the word embedding  and the output representation of BERT are located in the same vector space. Hence, we are able to recover the entity embedding from BERT's output representations to infuse their contextualized knowledge to the model.

To be specific, given a general or domain-specific corpus, we design to build the lookup table for entities that occurs in the downstream tasks on demand.  For an entity $e$, such as a  Wikidata entity or a proper noun entity, we construct its embedding $\textbf{E}(e)$  as follows:


\paragraph{Direction} A feasible method to add entity $e$ to the vocabulary of PLM is to optimize its embedding $\textbf{E}(e)$ for the MLM loss with other parameters frozen. 
We collect the sentences $S_{{e}}$ that contain entity $e$ and substitute it with $\texttt{[MASK]}$.  
The total influence of $\textbf{E}(e)$ to the MLM loss in $S_{{e}}$  can be formulated as:
\begin{equation}
\label{eq:proof}
\begin{split}
    \mathcal{L}(e) &= -\sum_{x_i\in S_{{e}}} \log\text{Pr}(e | \bm{r}_{x_i})\\
    &=   \sum_{{x_i} \in S_{{e}}} \log Z_{x_i} -  {{\textbf{E}}(e)}^T \sum_{{x_i} \in S_{{e}}} \bm{r}_{x_i}   ,
\end{split}
\end{equation}

where $Z_{x_i} =\sum_{w_j \in V\cup \{e\}} {\exp({{{\textbf{E}}(w_j)}^T \bm{r}_{x_i}}} )$,  $x_i$ is the replaced masked token for entity $e$ and $\bm{r}_{x_i}$ is the PLM's output representation of $x_i$.

Compared with the total impact of the entire vocabulary on $Z_{x_i}$, $\textbf{E}(e)$ has a much smaller impact. 
If we ignore the minor effect of $\textbf{E}(e)$ on $Z_{x_i}$,
the optimal solution of $\textbf{E}(e)$ for $\mathcal{L}(e)$ is proportional to $\sum_{{x_i} \in S_{{e}}} \bm{r}_{x_i}$. Hence, we set $\textbf{E}(e)$ as:
\begin{equation}
    \textbf{E}(e) =  C\cdot\sum_{x_i\in S_{{e}}} \bm{r}_{x_i},
    \label{eq:embdirection}
\end{equation}
where $C$  denotes the scaling factor.

Practically,  $\textbf{E}(e)$ also serves as the negative log-likelihood of other words' MLM loss~\cite{MutualInfo}. However, \citet{RepDegeneration} indicates that the gradient from such negative log-likelihood will push all  words to a uniformly negative direction, which weakens the quality of rare words' representation. Here, we ignore this negative term and  obtain the informative entity embedding from Eq.~\ref{eq:embdirection}.


\paragraph{Norm} We define $\bm{p}(e)$ as the position embedding for entity $e$. Since the layer normalization in Eq.~\ref{eq1}  makes the norm  $|\textbf{E}(e)+\bm{p}(e)|$ to ${D}^{\frac{1}{2}}$, we find that the norm $|\textbf{E}(e)|$ has little effect on the input feature of the  encoder in use. Therefore, we set the norm of all the entity embeddings as a constant $L$. Then, we evaluate the model with different $L$ on the unsupervised knowledge probe task and choose the best $L$ for those fine-tuning tasks.


\subsection{Infuse Entity Knowledge into PLMs}

Since the entity embedding we obtained and the original word embedding are both obtained from the masked language modeling objective, the entity can be regarded as a special input token. To infuse entity knowledge into PLMs, we apply a pair of bracket to enclose the constructed entity embedding and then insert it after the original entity's subwords. For example, the original  input,  

\emph{Most people with COVID-19 have a dry}  \texttt{[MASK]} \emph{they can feel in their chest.}
 
\noindent
becomes 

\emph{Most people with COVID-19 ({\textbf{COVID-19}})  have a dry}  \texttt{[MASK]} \emph{they can feel in their chest.}


\noindent
Here, the entity \emph{\textbf{COVID-19}} adopts our constructed entity embedding and other words use their original embedding. We simply convey the modified input to the PLM for encoding without any additional structures or parameters, to help the model predict \texttt{[MASK]} as \emph{cough}.

\paragraph{A note on entity links} In previous section, we hypothesize that we know the entity linking annotations for the involved string name. In practice, we can obtain the gold entity links provided by some datasets like FewRel 1.0. For the datasets where the linking annotations are not available, we employ a heuristic string matching for entity linking\footnote{Details are shown in the Appendix.}. 



\begin{table*}[!t]
\small
    \centering

 \resizebox{1.0\linewidth}{!}{
    \begin{tabular}{lc|cccc|cccc}
        \toprule
        \multicolumn{1}{l}{\multirow{2}{*}{\textbf{Model}}} & \multirow{2}{*}{\textbf{Ext. Pretrain}} &  \multicolumn{4}{c|}{\textbf{FewRel 1.0}} & \multicolumn{4}{c}{\textbf{FewRel 2.0}}\\
        & & \textbf{5-1} & \textbf{5-5} & \textbf{10-1} & \textbf{10-5} & \textbf{5-1} & \textbf{5-5} & \textbf{10-1} & \textbf{10-5}\\
        \midrule
     ERNIE$^\dagger$ & \checkmark & {92.7}$_{\pm {0.2}}$ & {97.9}$_{\pm {0.0}}$ & {87.7}$_{\pm {0.4}}$ & {96.1}$_{\pm {0.1}}$ & 66.4$_{\pm {1.6}}$ & 88.2$_{\pm {0.5}}$ & 51.2$_{\pm {0.7}}$ & 80.1$_{\pm {1.0}}$ \\   
       KEPLER & \checkmark & 90.8$_{\pm {0.1}}$ & 96.9$_{\pm {0.1}}$ & 85.1$_{\pm {0.1}}$ & 94.2$_{\pm {0.1}}$ & {74.0}$_{\pm {1.0}}$ &  {89.2}$_{\pm {0.2}}$ & {61.7}$_{\pm {0.1}}$& {82.1}$_{\pm {0.1}}$ \\
LUKE & \checkmark  & 91.8$_{\pm {0.4}}$ & {97.5}$_{\pm {0.1}}$ & 85.3$_{\pm {0.4}}$ & 95.3$_{\pm {0.1}}$ & 64.8$_{\pm {1.4}}$ & 89.2$_{\pm {0.2}}$ & 46.6$_{\pm {0.8}}$ & 80.5$_{\pm {0.5}}$ \\
        \midrule
RoBERTa & - & 90.4$_{\pm {0.3}}$ & 96.2$_{\pm {0.0}}$ & 84.2$_{\pm {0.5}}$ & 93.9$_{\pm {0.1}}$ & 71.2$_{\pm {2.1}}$ & 89.4$_{\pm {0.2}}$ & 53.3$_{\pm {0.8}}$ &  83.1$_{\pm {0.4}}$\\

        PELT & - & \textbf{92.7}$_{\pm {0.3}}$ & \textbf{97.5}$_{\pm {0.0}}$ &  \textbf{87.5}$_{\pm {0.3}}$ &  \textbf{95.4}$_{\pm {0.1}}$  & \textbf{75.0}$_{\pm {1.3}}$ & \textbf{92.1}$_{\pm {0.2}}$ &  \textbf{60.4}$_{\pm {1.1}}$  &  \textbf{85.6}$_{\pm {0.2}}$ \\ 
        \bottomrule
    \end{tabular}
    }

    \caption{The accuracy on the FewRel dataset. $N$-$K$ indicates the $N$-way $K$-shot configuration. Both of FewRel\,1.0 and FewRel\,2.0 are trained on the Wikipedia domain, and FewRel 2.0 is tested on the biomedical domain.  ERNIE$^\dagger$ has seen facts in the FewRel 1.0 test set  during pre-training. We report  standard deviations as subscripts.  
    }

    \label{tab:fewrel}
    

\end{table*}

\section{Experiment}

\subsection{Implementation Details}

We choose   RoBERTa$_{\small \texttt{Base}}$~\cite{RoBERTa}, a well-optimized PLM, as our baseline model and we equip it with our constructed entity embedding to obtain  the \Ourmodel model.  For the knowledge probe task, we further experiment with  another encoder-architecture model, uncased BERT$_{\small \texttt{Base}}$~\cite{BERT}, and an encoder-decoder-architecture model, BART$_{\small \texttt{Base}}$~\cite{BART}. 

We adopt Wikipedia and biomedical S2ORC~\cite{S2ORC} as the domain-specific corpora 
and split them into sentences with NLTK~\cite{NLTK}. For Wikipedia, we adopt a heuristic entity linking strategy with the help of hyperlink annotations. For the used FewRel 1.0 and Wiki80 datasets,  we directly use the annotated linking information.  
For other datasets, we link the given entity name through a simple string match. 
For each necessary entity, we first extract up to 256 sentences containing the entity from the corpora.  We adopt Wikipedia as the domain-specific corpus for FewRel 1.0, Wiki80 and LAMA,  and we adopt S2ORC as the  domain-specific corpus for  FewRel 2.0. 
After that, we construct the entity embedding according to Section~\ref{sec:construct}. 

We search the norm of entity embedding $L$ among 1-10 on the knowledge probe task. We find $L=7,10,3$ performs a bit better for RoBERTa, BERT and BART respectively. 
In the fine-tuning process, we freeze the constructed embeddings  as an lookup table with the corresponding norm. After that, we run all the fine-tuning experiments with 5 different seeds and report the average score. 

\subsection{Baselines}

We select three of the most representative  entity-aware baselines, which adopt an external entity embedding, an entity-related pre-training task, or a trainable entity embedding: (1) \textbf{ERNIE}~\cite{ERNIE} involves the entity embedding learned from Wikidata relation~\cite{TransE}. We adopt the RoBERTa version of ERNIE provided by \citet{KEPLER};  
(2) \textbf{KEPLER}~\cite{KEPLER} encodes textual entity description into entity embedding and learns fact triples and  language modeling  simultaneously; 
(3) \textbf{LUKE}~\cite{LUKE} learns a trainable entity embedding to help the model predict masked tokens and masked entities in the sentences.


\begin{table}[!t]
\small
    \centering
    \begin{tabular}{l|ccc}
        \toprule
        \textbf{Model}& \textbf{1\%} & \textbf{10\%} & \textbf{100\%}  \\
        \midrule
ERNIE & {66.4}$_{\pm {0.4}}$ & {87.7}$_{\pm {0.2}}$ & {93.4}$_{\pm {0.1}}$ \\
 KEPLER& 62.3$_{\pm {1.0}}$& 85.4$_{\pm {0.2}}$& 91.7$_{\pm {0.1}}$ \\
LUKE & 63.1$_{\pm {1.0}}$& 86.9$_{\pm {0.4}}$& 92.9$_{\pm {0.1}}$\\
\midrule
RoBERTa & 59.8$_{\pm {1.7}}$ & 85.7$_{\pm {0.2}}$ & 91.7$_{\pm {0.1}}$ \\
PELT &  \textbf{65.6}$_{\pm {1.0}}$ & \textbf{88.3}$_{\pm {0.3}}$ & \textbf{93.4}$_{\pm {0.1}}$  \\
        \bottomrule
    \end{tabular}
 \caption{The accuracy on the test set of Wiki80. 1\%/\,10\% indicate using  1\%/\,10\% supervised training data respectively.  
 }
 
 \label{tab:re}

\end{table}

\subsection{Relation Classification}

Relation Classification (RC) aims to predict the relationship between two entities in a given text. We evaluate the models on two scenarios, the few-shot setting and the full-data setting. 

The few-shot setting focuses on   long-tail relations without sufficient training instances. We evaluate models on FewRel 1.0~\cite{Fewrel} and FewRel 2.0~\cite{Fewrel2.0}. FewRel 1.0 contains instances with Wikidata facts and 
FewRel 2.0 involves a  biomedical-domain test set to examine the ability of domain adaptation. In the $N$-way $K$-shot setting,  models are required to categorize the query as one of the existing $N$ relations, each of which contains $K$ supporting samples. We choose the state-of-the-art few-shot framework Proto~\cite{Proto} with different PLM encoders for evaluation.  
For the full-data setting, we evaluate models on the  Wiki80, which contains 80 relation types from Wikidata. We also add 1\% and 10\% settings, meaning using only 1\% / 10\% data of the training sets.

As shown in Table~\ref{tab:fewrel} and Table~\ref{tab:re}, on FewRel 1.0 and Wiki80 in Wikipedia domain, RoBERTa with PELT beats the RoBERTa model by a large margin\,(e.g. +3.3\% on 10way-1shot), and it even achieves comparable performance with ERNIE, which has access to the knowledge graph.  
Our model also gains huge improvements on FewRel 2.0 in the biomedical domain\,(e.g. +$7.1\%$ on 10way-1shot), while the entity-aware baselines have little advance in most settings. Compared with most existing entity-aware PLMs which merely obtain domain-specific knowledge in the pre-training phase, our proposed pluggable entity lookup table can dynamically update the models' knowledge from the out-of-domain corpus on demand.





\subsection{Knowledge Probe}
\begin{table}[t!]
\small
\centering
\begin{tabular}{l|cc|cc}
\toprule
\multirow{2}{*}{\textbf{Model}} & \multicolumn{2}{c|}{\textbf{LAMA}}   & \multicolumn{2}{c}{\textbf{LAMA-UHN}} \\ \cmidrule{2-5} 
& \textbf{G-RE} & \textbf{T-REx}    &  \textbf{G-RE}  & \textbf{T-REx}  \\ 
\midrule
ERNIE            & 10.0         & 24.9    & 5.9           & 19.4      \\ 
KEPLER            & 5.5         & 23.4    & 2.5           & 15.4      \\ 
LUKE & 3.8 & 32.0 &  2.0 & 25.3 \\
\midrule
RoBERTa            & 5.4         & 24.7    & 2.2           & 17.0      \\ 
PELT & \textbf{6.4} & \textbf{27.5}   & \textbf{2.8} & \textbf{19.3} \\
\midrule
\midrule
BERT & \textbf{13.9} & 34.9 & 8.8 & 26.8 \\
BERT-PELT & 13.3& \textbf{40.7} &\textbf{8.9} & \textbf{34.5}\\
\midrule
BART & 5.1 & 21.2 & 1.3 & 12.0 \\
BART-PELT & \textbf{6.9}& \textbf{24.4}& \textbf{2.1} & \textbf{14.9}\\
\bottomrule
\end{tabular}
 \caption{Mean P@1 on the knowledge probe benchmark. G-RE: Google-RE. }
 \label{tab:lama}
\end{table}

We conduct experiments on a widely-used knowledge probe dataset, LAMA~\cite{LAMA}. It applies cloze-style questions to examine PLMs’ ability on recalling  facts from their parameters.   For example, given a question template \emph{Paris is the capital of} \texttt{[MASK]}, PLMs are required to predict the masked token properly.  In this paper, we not only use  Gooogle-RE and  T-REx~\cite{T-REX} which focus on factual knowledge, but also evaluate models on LAMA-UHN~\cite{LAMA-UHN} which filters out the easy questionable templates. 

As shown in Table~\ref{tab:lama}, without any pre-training, the PELT model can directly absorb the entity knowledge from the extended input sequence to recall more factual knowledge, which demonstrates that the entity embeddings we constructed are compatible with original word embeddings.   We also find that our method can also bring huge improvements to both  BERT  and BART in the knowledge probe task, which proves our method’s generalization on different-architecture PLMs.


\paragraph{Effect of Entity Frequency}
Table~\ref{tab:freq} shows the P@1 results with respect to the entity   frequency. While RoBERTa performs worse on rare entities than frequent entities, \Ourmodel brings a substantial improvement on rare entities, i.e., near 3.8  mean P@1 gains on entities that occur less than 50 times.

\begin{table}[!t]
\small
    \centering
    \begin{tabular}{l|cccc}
        \toprule
\textbf{Model} & [0,10) & {[10,50)} & [50,100)   & [100,+) \\
\midrule
RoBERTa & 18.1 & 21.1 & 25.8 & 26.1 \\
PELT & \textbf{21.9} & \textbf{24.8} & \textbf{29.0} & \textbf{28.7} \\
        \bottomrule
    \end{tabular}
    \caption{Mean P@1 on T-Rex with respect to the subject entity's frequency  in Wikipedia. }
    \label{tab:freq}

\end{table}

\section{Conclusion}
In this paper, we propose \Ourmodel, a flexible entity lookup table, to incorporate up-to-date knowledge into PLMs. By constructing entity embeddings on demand, PLMs with \Ourmodel can recall rich factual knowledge to help downstream tasks.

\section*{Acknowledgement}
This work is supported by the National Key R\&D Program of China (No.\,2020AAA0106502), Institute Guo Qiang at Tsinghua University, and International Innovation Center of Tsinghua University, Shanghai, China. We thank Zhengyan Zhang and other members of THUNLP for
their helpful discussion and feedback. Deming Ye conducted the experiments. Deming Ye, Yankai Lin, Xiaojun Xie and Peng Li wrote the paper. Maosong Sun and Zhiyuan Liu provided valuable advices to the research. 



\bibliography{anthology,custom}

\begin{thebibliography}{29}
\expandafter\ifx\csname natexlab\endcsname\relax\def\natexlab#1{#1}\fi

\bibitem[{Ba et~al.(2016)Ba, Kiros, and Hinton}]{LayerNorm}
Lei~Jimmy Ba, Jamie~Ryan Kiros, and Geoffrey~E. Hinton. 2016.
\newblock \href {http://arxiv.org/abs/1607.06450} {Layer normalization}.
\newblock \emph{CoRR}, abs/1607.06450.

\bibitem[{Bordes et~al.(2013)Bordes, Usunier, Garc{\'{\i}}a{-}Dur{\'{a}}n,
  Weston, and Yakhnenko}]{TransE}
Antoine Bordes, Nicolas Usunier, Alberto Garc{\'{\i}}a{-}Dur{\'{a}}n, Jason
  Weston, and Oksana Yakhnenko. 2013.
\newblock \href
  {https://proceedings.neurips.cc/paper/2013/hash/1cecc7a77928ca8133fa24680a88d2f9-Abstract.html}
  {Translating embeddings for modeling multi-relational data}.
\newblock In \emph{Advances in Neural Information Processing Systems 26: 27th
  Annual Conference on Neural Information Processing Systems 2013. Proceedings
  of a meeting held December 5-8, 2013, Lake Tahoe, Nevada, United States},
  pages 2787--2795.

\bibitem[{Devlin et~al.(2019)Devlin, Chang, Lee, and Toutanova}]{BERT}
Jacob Devlin, Ming-Wei Chang, Kenton Lee, and Kristina Toutanova. 2019.
\newblock \href {https://doi.org/10.18653/v1/N19-1423} {{BERT}: Pre-training of
  deep bidirectional transformers for language understanding}.
\newblock In \emph{Proceedings of the 2019 Conference of the North {A}merican
  Chapter of the Association for Computational Linguistics: Human Language
  Technologies, Volume 1 (Long and Short Papers)}, pages 4171--4186,
  Minneapolis, Minnesota. Association for Computational Linguistics.

\bibitem[{ElSahar et~al.(2018)ElSahar, Vougiouklis, Remaci, Gravier, Hare,
  Laforest, and Simperl}]{T-REX}
Hady ElSahar, Pavlos Vougiouklis, Arslen Remaci, Christophe Gravier,
  Jonathon~S. Hare, Fr{\'{e}}d{\'{e}}rique Laforest, and Elena Simperl. 2018.
\newblock \href
  {http://www.lrec-conf.org/proceedings/lrec2018/summaries/632.html} {T-rex:
  {A} large scale alignment of natural language with knowledge base triples}.
\newblock In \emph{Proceedings of the Eleventh International Conference on
  Language Resources and Evaluation, {LREC} 2018, Miyazaki, Japan, May 7-12,
  2018}. European Language Resources Association {(ELRA)}.

\bibitem[{F{\'e}vry et~al.(2020)F{\'e}vry, Baldini~Soares, FitzGerald, Choi,
  and Kwiatkowski}]{EAE}
Thibault F{\'e}vry, Livio Baldini~Soares, Nicholas FitzGerald, Eunsol Choi, and
  Tom Kwiatkowski. 2020.
\newblock \href {https://doi.org/10.18653/v1/2020.emnlp-main.400} {Entities as
  experts: Sparse memory access with entity supervision}.
\newblock In \emph{Proceedings of the 2020 Conference on Empirical Methods in
  Natural Language Processing (EMNLP)}, pages 4937--4951, Online. Association
  for Computational Linguistics.

\bibitem[{Gao et~al.(2019{\natexlab{a}})Gao, He, Tan, Qin, Wang, and
  Liu}]{RepDegeneration}
Jun Gao, Di~He, Xu~Tan, Tao Qin, Liwei Wang, and Tie{-}Yan Liu.
  2019{\natexlab{a}}.
\newblock \href {https://openreview.net/forum?id=SkEYojRqtm} {Representation
  degeneration problem in training natural language generation models}.
\newblock In \emph{7th International Conference on Learning Representations,
  {ICLR} 2019, New Orleans, LA, USA, May 6-9, 2019}. OpenReview.net.

\bibitem[{Gao et~al.(2019{\natexlab{b}})Gao, Han, Zhu, Liu, Li, Sun, and
  Zhou}]{Fewrel2.0}
Tianyu Gao, Xu~Han, Hao Zhu, Zhiyuan Liu, Peng Li, Maosong Sun, and Jie Zhou.
  2019{\natexlab{b}}.
\newblock \href {https://doi.org/10.18653/v1/D19-1649} {{F}ew{R}el 2.0: Towards
  more challenging few-shot relation classification}.
\newblock In \emph{Proceedings of the 2019 Conference on Empirical Methods in
  Natural Language Processing and the 9th International Joint Conference on
  Natural Language Processing (EMNLP-IJCNLP)}, pages 6250--6255, Hong Kong,
  China. Association for Computational Linguistics.

\bibitem[{Han et~al.(2018)Han, Zhu, Yu, Wang, Yao, Liu, and Sun}]{Fewrel}
Xu~Han, Hao Zhu, Pengfei Yu, Ziyun Wang, Yuan Yao, Zhiyuan Liu, and Maosong
  Sun. 2018.
\newblock \href {https://doi.org/10.18653/v1/D18-1514} {{F}ew{R}el: A
  large-scale supervised few-shot relation classification dataset with
  state-of-the-art evaluation}.
\newblock In \emph{Proceedings of the 2018 Conference on Empirical Methods in
  Natural Language Processing}, pages 4803--4809, Brussels, Belgium.
  Association for Computational Linguistics.

\bibitem[{Kingma and Ba(2015)}]{Adam}
Diederik~P. Kingma and Jimmy Ba. 2015.
\newblock \href {http://arxiv.org/abs/1412.6980} {Adam: {A} method for
  stochastic optimization}.
\newblock In \emph{3rd International Conference on Learning Representations,
  {ICLR} 2015, San Diego, CA, USA, May 7-9, 2015, Conference Track
  Proceedings}.

\bibitem[{Kong et~al.(2020)Kong, de~Masson~d'Autume, Yu, Ling, Dai, and
  Yogatama}]{MutualInfo}
Lingpeng Kong, Cyprien de~Masson~d'Autume, Lei Yu, Wang Ling, Zihang Dai, and
  Dani Yogatama. 2020.
\newblock \href {https://openreview.net/forum?id=Syx79eBKwr} {A mutual
  information maximization perspective of language representation learning}.
\newblock In \emph{8th International Conference on Learning Representations,
  {ICLR} 2020, Addis Ababa, Ethiopia, April 26-30, 2020}. OpenReview.net.

\bibitem[{Lewis et~al.(2020)Lewis, Liu, Goyal, Ghazvininejad, Mohamed, Levy,
  Stoyanov, and Zettlemoyer}]{BART}
Mike Lewis, Yinhan Liu, Naman Goyal, Marjan Ghazvininejad, Abdelrahman Mohamed,
  Omer Levy, Veselin Stoyanov, and Luke Zettlemoyer. 2020.
\newblock \href {https://doi.org/10.18653/v1/2020.acl-main.703} {{BART:}
  denoising sequence-to-sequence pre-training for natural language generation,
  translation, and comprehension}.
\newblock In \emph{Proceedings of the 58th Annual Meeting of the Association
  for Computational Linguistics, {ACL} 2020, Online, July 5-10, 2020}, pages
  7871--7880. Association for Computational Linguistics.

\bibitem[{Liu et~al.(2019)Liu, Ott, Goyal, Du, Joshi, Chen, Levy, Lewis,
  Zettlemoyer, and Stoyanov}]{RoBERTa}
Yinhan Liu, Myle Ott, Naman Goyal, Jingfei Du, Mandar Joshi, Danqi Chen, Omer
  Levy, Mike Lewis, Luke Zettlemoyer, and Veselin Stoyanov. 2019.
\newblock {RoBERTa}: {A} robustly optimized {BERT} pretraining approach.
\newblock \emph{CoRR}, abs/1907.11692.

\bibitem[{Lo et~al.(2020)Lo, Wang, Neumann, Kinney, and Weld}]{S2ORC}
Kyle Lo, Lucy~Lu Wang, Mark Neumann, Rodney Kinney, and Daniel Weld. 2020.
\newblock \href {https://doi.org/10.18653/v1/2020.acl-main.447} {{S}2{ORC}: The
  semantic scholar open research corpus}.
\newblock In \emph{Proceedings of the 58th Annual Meeting of the Association
  for Computational Linguistics}, pages 4969--4983, Online. Association for
  Computational Linguistics.

\bibitem[{Peters et~al.(2019)Peters, Neumann, Logan, Schwartz, Joshi, Singh,
  and Smith}]{KnowBERT}
Matthew~E. Peters, Mark Neumann, Robert Logan, Roy Schwartz, Vidur Joshi,
  Sameer Singh, and Noah~A. Smith. 2019.
\newblock \href {https://doi.org/10.18653/v1/D19-1005} {Knowledge enhanced
  contextual word representations}.
\newblock In \emph{Proceedings of the 2019 Conference on Empirical Methods in
  Natural Language Processing and the 9th International Joint Conference on
  Natural Language Processing (EMNLP-IJCNLP)}, pages 43--54, Hong Kong, China.
  Association for Computational Linguistics.

\bibitem[{Petroni et~al.(2019)Petroni, Rockt{\"a}schel, Riedel, Lewis, Bakhtin,
  Wu, and Miller}]{LAMA}
Fabio Petroni, Tim Rockt{\"a}schel, Sebastian Riedel, Patrick Lewis, Anton
  Bakhtin, Yuxiang Wu, and Alexander Miller. 2019.
\newblock \href {https://doi.org/10.18653/v1/D19-1250} {Language models as
  knowledge bases?}
\newblock In \emph{Proceedings of the 2019 Conference on Empirical Methods in
  Natural Language Processing and the 9th International Joint Conference on
  Natural Language Processing (EMNLP-IJCNLP)}, pages 2463--2473, Hong Kong,
  China. Association for Computational Linguistics.

\bibitem[{P{\"{o}}rner et~al.(2020)P{\"{o}}rner, Waltinger, and
  Sch{\"{u}}tze}]{LAMA-UHN}
Nina P{\"{o}}rner, Ulli Waltinger, and Hinrich Sch{\"{u}}tze. 2020.
\newblock \href {https://doi.org/10.18653/v1/2020.findings-emnlp.71} {{E-BERT:}
  efficient-yet-effective entity embeddings for {BERT}}.
\newblock In \emph{Findings of the Association for Computational Linguistics:
  {EMNLP} 2020, Online Event, 16-20 November 2020}, volume {EMNLP} 2020 of
  \emph{Findings of {ACL}}, pages 803--818. Association for Computational
  Linguistics.

\bibitem[{Roberts et~al.(2020)Roberts, Raffel, and Shazeer}]{T5closedbookqa}
Adam Roberts, Colin Raffel, and Noam Shazeer. 2020.
\newblock \href {https://doi.org/10.18653/v1/2020.emnlp-main.437} {How much
  knowledge can you pack into the parameters of a language model?}
\newblock In \emph{Proceedings of the 2020 Conference on Empirical Methods in
  Natural Language Processing, {EMNLP} 2020, Online, November 16-20, 2020},
  pages 5418--5426. Association for Computational Linguistics.

\bibitem[{Rogers et~al.(2020)Rogers, Kovaleva, and Rumshisky}]{BERTology}
Anna Rogers, Olga Kovaleva, and Anna Rumshisky. 2020.
\newblock \href {https://transacl.org/ojs/index.php/tacl/article/view/2257} {A
  primer in bertology: What we know about how {BERT} works}.
\newblock \emph{Trans. Assoc. Comput. Linguistics}, 8:842--866.

\bibitem[{Snell et~al.(2017)Snell, Swersky, and Zemel}]{Proto}
Jake Snell, Kevin Swersky, and Richard~S. Zemel. 2017.
\newblock \href
  {https://proceedings.neurips.cc/paper/2017/hash/cb8da6767461f2812ae4290eac7cbc42-Abstract.html}
  {Prototypical networks for few-shot learning}.
\newblock In \emph{Advances in Neural Information Processing Systems 30: Annual
  Conference on Neural Information Processing Systems 2017, December 4-9, 2017,
  Long Beach, CA, {USA}}, pages 4077--4087.

\bibitem[{Sun et~al.(2020)Sun, Shao, Qiu, Guo, Hu, Huang, and Zhang}]{CoLAKE}
Tianxiang Sun, Yunfan Shao, Xipeng Qiu, Qipeng Guo, Yaru Hu, Xuanjing Huang,
  and Zheng Zhang. 2020.
\newblock \href {https://doi.org/10.18653/v1/2020.coling-main.327} {{C}o{LAKE}:
  Contextualized language and knowledge embedding}.
\newblock In \emph{Proceedings of the 28th International Conference on
  Computational Linguistics}, pages 3660--3670, Barcelona, Spain (Online).
  International Committee on Computational Linguistics.

\bibitem[{Tenney et~al.(2019)Tenney, Xia, Chen, Wang, Poliak, McCoy, Kim,
  Durme, Bowman, Das, and Pavlick}]{Prob}
Ian Tenney, Patrick Xia, Berlin Chen, Alex Wang, Adam Poliak, R.~Thomas McCoy,
  Najoung Kim, Benjamin~Van Durme, Samuel~R. Bowman, Dipanjan Das, and Ellie
  Pavlick. 2019.
\newblock \href {https://openreview.net/forum?id=SJzSgnRcKX} {What do you learn
  from context? probing for sentence structure in contextualized word
  representations}.
\newblock In \emph{7th International Conference on Learning Representations,
  {ICLR} 2019, New Orleans, LA, USA, May 6-9, 2019}. OpenReview.net.

\bibitem[{Wang et~al.(2021{\natexlab{a}})Wang, Liu, and
  Zhang}]{canclosedbookqa}
Cunxiang Wang, Pai Liu, and Yue Zhang. 2021{\natexlab{a}}.
\newblock \href {https://doi.org/10.18653/v1/2021.acl-long.251} {Can generative
  pre-trained language models serve as knowledge bases for closed-book qa?}
\newblock In \emph{Proceedings of the 59th Annual Meeting of the Association
  for Computational Linguistics and the 11th International Joint Conference on
  Natural Language Processing, {ACL/IJCNLP} 2021, (Volume 1: Long Papers),
  Virtual Event, August 1-6, 2021}, pages 3241--3251. Association for
  Computational Linguistics.

\bibitem[{Wang et~al.(2020)Wang, Tang, Duan, Wei, Huang, Ji, Cao, Jiang, and
  Zhou}]{K-Adapter}
Ruize Wang, Duyu Tang, Nan Duan, Zhongyu Wei, Xuanjing Huang, Jianshu Ji,
  Guihong Cao, Daxin Jiang, and Ming Zhou. 2020.
\newblock \href {http://arxiv.org/abs/2002.01808} {K-adapter: Infusing
  knowledge into pre-trained models with adapters}.
\newblock \emph{CoRR}, abs/2002.01808.

\bibitem[{Wang et~al.(2021{\natexlab{b}})Wang, Gao, Zhu, Zhang, Liu, Li, and
  Tang}]{KEPLER}
Xiaozhi Wang, Tianyu Gao, Zhaocheng Zhu, Zhengyan Zhang, Zhiyuan Liu, Juanzi
  Li, and Jian Tang. 2021{\natexlab{b}}.
\newblock \href {https://transacl.org/ojs/index.php/tacl/article/view/2447}
  {{KEPLER:} {A} unified model for knowledge embedding and pre-trained language
  representation}.
\newblock \emph{Trans. Assoc. Comput. Linguistics}, 9:176--194.

\bibitem[{Xiong et~al.(2020)Xiong, Du, Wang, and Stoyanov}]{WKLM}
Wenhan Xiong, Jingfei Du, William~Yang Wang, and Veselin Stoyanov. 2020.
\newblock \href {https://openreview.net/forum?id=BJlzm64tDH} {Pretrained
  encyclopedia: Weakly supervised knowledge-pretrained language model}.
\newblock In \emph{8th International Conference on Learning Representations,
  {ICLR} 2020, Addis Ababa, Ethiopia, April 26-30, 2020}. OpenReview.net.

\bibitem[{Xue(2011)}]{NLTK}
Nianwen Xue. 2011.
\newblock \href {https://doi.org/10.1017/S1351324910000306} {Steven bird, evan
  klein and edward loper. \emph{Natural Language Processing with Python}.
  o'reilly media, inc 2009. {ISBN:} 978-0-596-51649-9}.
\newblock \emph{Nat. Lang. Eng.}, 17(3):419--424.

\bibitem[{Yamada et~al.(2020)Yamada, Asai, Shindo, Takeda, and
  Matsumoto}]{LUKE}
Ikuya Yamada, Akari Asai, Hiroyuki Shindo, Hideaki Takeda, and Yuji Matsumoto.
  2020.
\newblock \href {https://doi.org/10.18653/v1/2020.emnlp-main.523} {{LUKE}: Deep
  contextualized entity representations with entity-aware self-attention}.
\newblock In \emph{Proceedings of the 2020 Conference on Empirical Methods in
  Natural Language Processing (EMNLP)}, pages 6442--6454, Online. Association
  for Computational Linguistics.

\bibitem[{Yamada et~al.(2016)Yamada, Shindo, Takeda, and
  Takefuji}]{Wikipedia2Vec}
Ikuya Yamada, Hiroyuki Shindo, Hideaki Takeda, and Yoshiyasu Takefuji. 2016.
\newblock \href {https://doi.org/10.18653/v1/k16-1025} {Joint learning of the
  embedding of words and entities for named entity disambiguation}.
\newblock In \emph{Proceedings of the 20th {SIGNLL} Conference on Computational
  Natural Language Learning, CoNLL 2016, Berlin, Germany, August 11-12, 2016},
  pages 250--259. {ACL}.

\bibitem[{Zhang et~al.(2019)Zhang, Han, Liu, Jiang, Sun, and Liu}]{ERNIE}
Zhengyan Zhang, Xu~Han, Zhiyuan Liu, Xin Jiang, Maosong Sun, and Qun Liu. 2019.
\newblock \href {https://doi.org/10.18653/v1/P19-1139} {{ERNIE}: Enhanced
  language representation with informative entities}.
\newblock In \emph{Proceedings of the 57th Annual Meeting of the Association
  for Computational Linguistics}, pages 1441--1451, Florence, Italy.
  Association for Computational Linguistics.

\end{thebibliography}
\bibliographystyle{acl_natbib}

\appendix

\section{Heuristic String Matching for Entity Linking}

For the Wikipedia, we first create a mapping from the anchor texts with hyperlinks to their referent Wikipedia pages. After that, We employ a heuristic string matching to link other potential entities to their pages.
 
For preparation, we collect the aliases of the entity from the redirect page of Wikipedia and the relation between entities from the hyperlink.  Then, we apply spaCy~\footnote{\url{https://spacy.io/}} to recognize the entity name in the text. An entity name in the text may refer to multiple entities of the same alias. We utilize the relation of the linked entity page to maintain an available entity page set for entity disambiguation .

\begin{algorithm}
\begin{algorithmic}
\STATE $S \Leftarrow \{$ the linked entity page in anchor text$\}$
\STATE $E\Leftarrow\{$ potential entity name in text$\}$
\REPEAT 
\STATE $S'\Leftarrow\{$ the neighbor entity pages that have hyperlink or Wikidata relation with pages in $S\}$
\STATE $E'\Leftarrow\{e | e\in E$ and $e$ can be uniquely linked to entity page in $S'$ by string matching $\}$
\STATE $E \Leftarrow E - E'$
\STATE $S \Leftarrow E'$
\UNTIL{$S=\phi$}
\end{algorithmic}
\caption{Heuristic string matching for entity disambiguation}
\label{alg:A}
\end{algorithm}
Details of the heuristic string matching are shown in Algorithm~\ref{alg:A}, we match the entity name to surrounding entity page of the current page as close as possible. e will release  all the source code and models with the pre-processed Wikipedia dataset.

For other datases, we adopt a simple string matching for entity linking.

\section{Training Configuration}


We train all the models with Adam optimizer~\cite{Adam}, 10\% warming up steps and maximum 128 input tokens. Detailed training  hyper-parameters are shown in Table~\ref{tab:Hyperparameters}. 

We run all the experiments with 5 different seeds (42, 43, 44, 45, 46)  and report the average score with the standard deviation. In the 1\% and 10\% settings' experiments for Wiki80, we train the model with 10-25 times epochs as that of the 100\% setting's experiment. 

For FewRel, we search the batch size among [4,8,32] and search the training step in [1500, 2000, 2500]. We evaluate models every 250 on validation and save the model with best performance for testing. With our hyper-parameter tuning, the results of baselines in FewRel significantly outperforms that reported by KEPLER~\cite{KEPLER}.

\begin{table}[ht]
\centering
\small
\begin{tabular}{l | c c c c } 
\toprule
Dataset &  Epoch & Train Step  & BSZ & LR \\
\midrule
Wiki80 & 5& -& 32 & 3e-5 \\
FewRel\,1.0 & -& 1500 & 32 &  2e-5 \\
FewRel\,2.0 & -& 1500 & 32 & 2e-5\\
\bottomrule
\end{tabular}
\caption{Training Hyper-parameters. BSZ: Batch size; LR: Learning rate.}
\label{tab:Hyperparameters}
\end{table}



\end{document}